\documentclass[draft]{agujournal2019}
\usepackage[utf8]{inputenc}
\usepackage[T1]{fontenc}
\usepackage[utf8]{inputenc}
\usepackage{booktabs}
\usepackage{multirow}
\usepackage{enumitem}
\usepackage{url} 
\usepackage{lineno}
\usepackage[inline]{trackchanges} 
\usepackage{soul}
\usepackage{fancyhdr}
\draftfalse

\begin{document}


\title{mloz: A Highly Efficient Machine Learning-Based Ozone Parameterization for Climate Sensitivity Simulations}

%
%

\authors{Yiling Ma\affil{1}, Nathan Luke Abraham\affil{2,3}, Stefan Versick\affil{1}, Roland Ruhnke\affil{1}, Andrea Schneidereit\affil{4}, Ulrike Niemeier\affil{5}, Felix Back\affil{6}, Peter Braesicke\affil{4}, Peer Nowack\affil{6,1}}

\affiliation{1}{Institute of Meteorology and Climate Research Atmospheric Trace Gases and Remote Sensing, Karlsruhe Institute of Technology (KIT), Eggenstein-Leopoldshafen, Germany}
\affiliation{2}{Yusuf Hamied Department of Chemistry, University of Cambridge, Cambridge, UK}
\affiliation{3}{National Centre for Atmospheric Science, UK}
\affiliation{4}{Deutscher Wetterdienst, Offenbach am Main, Germany}
\affiliation{5}{Max Planck Institute for Meteorology, Hamburg, Germany}
\affiliation{6}{Institute of Theoretical Informatics, Karlsruhe Institute of Technology (KIT), Karlsruhe, Germany}

\correspondingauthor{Yiling Ma}{yiling.ma@kit.edu}


\begin{keypoints}
\item mloz is introduced and implemented to interactively model daily ozone variability and trends in standard climate sensitivity simulations
\item mloz produces accurate three-dimensional ozone fields in multi-decadal simulations at hardly any added computational cost
\item The parameterization is demonstrated to be transferable across climate models by successful tests in ICON after training on UKESM data
\end{keypoints}

%
%

\begin{abstract}
Atmospheric ozone is a crucial absorber of solar radiation and an important greenhouse gas. However, most climate models participating in the Coupled Model Intercomparison Project (CMIP) still lack an interactive representation of ozone due to the high computational costs of atmospheric chemistry schemes. Here, we introduce a machine learning parameterization (mloz) to interactively model daily ozone variability and trends across the troposphere and stratosphere in standard climate sensitivity simulations, including two-way interactions of ozone with the Quasi-Biennial Oscillation. We demonstrate its high fidelity on decadal timescales and its flexible use online across two different climate models -- the UK Earth System Model (UKESM) and the German ICOsahedral Nonhydrostatic (ICON) model. With atmospheric temperature profile information as the only input, mloz produces stable ozone predictions $\sim$31 times faster than the chemistry scheme in UKESM, contributing less than 4\% of the respective total climate model runtimes. In particular, we also demonstrate its transferability to different climate models without chemistry schemes by transferring the parameterization from UKESM to ICON. This highlights mloz’s potential for widespread adoption in CMIP-level climate models that lack interactive chemistry for future climate change assessments, particularly when focusing on climate sensitivity simulations, where ozone trends and variability are known to significantly modulate atmospheric feedback processes.
\end{abstract}

\section*{Plain Language Summary}
Ozone plays an important role in the climate system by acting as a greenhouse gas and by absorbing solar radiation, which ultimately affects atmospheric temperatures and circulation patterns from the stratosphere down to Earth's surface. However, most climate models used in major international assessments still rely on fixed climatological ozone fields that cannot respond to changing conditions within climate change simulations. This is mainly because simulating ozone interactively requires complex chemistry and transport calculations, which are highly computationally expensive. We here present mloz -- the first fully interactive machine learning parameterization to represent ozone variability and trends in climate models. Using temperature profiles, mloz predict daily ozone concentrations across the atmosphere and captures ozone-climate feedback under changing climate conditions at almost no extra computational cost. We tested mloz in two major climate models (UKESM and ICON) and found it works reliably in long simulations. Importantly, mloz trained on UKESM data also performed well in ICON, showing it can be transferred across different models. This approach allows climate models that lack atmospheric chemistry schemes to include a fast and interactive ozone representation, supporting broader use of interactive chemistry in Earth system modeling and helping policymakers to better understand how climate may evolve.

%
%

\section{Introduction}
Atmospheric ozone plays multiple vital roles in the Earth system and has significant impacts on human health. Most ozone is located in the stratosphere, forming the ozone layer, where it absorbs highly harmful solar ultraviolet (UV) radiation, thus protecting life on Earth~\cite{WMO2022}. In contrast, ozone in the troposphere acts as an air pollutant that exaggerates, among others, respiratory diseases and causes plant damage~\cite{Donzelli2024}. Ozone is also an important greenhouse gas (GHG) and its historical increases in the troposphere have already led to substantial radiative forcing~\cite{IPCC2021Chapter6,WMO2022}. Moreover, the feedback of tropospheric and stratospheric ozone plays an important role in modulating surface warming and the atmospheric dynamics response to GHGs increases~\cite{jonsson2004doubled,nowack2015large,Dietmuller2014,chiodo2019jc,Muthers2014gmd}. 

The main mechanisms governing ozone formation and loss are distinct between the stratosphere and the troposphere~\cite{monks2015tropospheric,Solomon1999}. In the stratosphere, ozone concentrations are primarily controlled by photochemical reactions following the Chapman cycle~\cite{chapman1930}. These are modulated by catalytic ozone loss cycles, in particular those involving hydrogen oxides, nitrogen oxides (${\mathrm{NO}_\mathrm{x}}$), and halogen radicals~\cite{lary1997catalytic}. Ozone loss through halogen species has been substantially enhanced by anthropogenic emissions of halogenated ozone-depleting substances, which has led to the formation of Antarctic ozone holes~\cite{WMO2022}. In addition to chemistry, dynamical processes -- particularly stratospheric transport as part of the Brewer-Dobson Circulation (BDC) -- play a crucial role in shaping ozone distributions, especially in the lower stratosphere where ozone lifetimes are longer~\cite{plumb2002stratospheric}. Ozone variability is also influenced by natural climate variability such as the El Niño–Southern Oscillation (ENSO)~\cite{benito2022driving}, the Quasi-Biennial Oscillation (QBO)~\cite{zawodny1991stratospheric}, volcanic eruptions~\cite{Solomon2016}, and solar activity~\cite{haigh1994role}. Looking ahead, future stratospheric ozone is projected to evolve associated with GHG emission trajectories~\cite{keeble2021evaluating}, as GHG forcings will modulate ozone concentrations through their effects on atmospheric transport, background environmental conditions (e.g., temperature and humidity), and photolysis rates~\cite{Fu_2019,Meul2014acp,molina1974stratospheric}. In particular, background temperature changes induce ozone anomalies via the temperature dependency of photochemical and catalytic reactions~\cite{Hocke2023atmo}. Tropospheric ozone, in turn, is primarily controlled by pre-cursor emissions -- in particular of volatile organic compounds (VOCs) in the presence of ${\mathrm{NO}_\mathrm{x}}$ -- and subsequent chemical production and loss cycles, as well as dry deposition at the surface, while also being subject to long-range transport~\cite{lu2019meteorology}. Therefore, an explicit representation of ozone requires to resolve many processes governing ozone variations, involving radiative, dynamical, and chemical coupling between ozone and the climate system on various timescales.

In addition to being affected by background temperature conditions, ozone in turn leaves a characteristic imprint on the atmospheric temperature profile. By absorbing solar energy for photolysis, ozone and its production process heats up the upper stratosphere by more than 20~K~\cite{matsumi2003photolysis}. Its short-wave and long-wave radiative heating is substantial throughout the atmosphere and ultimately also affects surface temperatures~\cite{WMO2022}, with the relative importance of short-wave and long-wave ozone forcing depending strongly on latitude and altitude. Since climate change is expected to substantially affect future ozone concentrations, the subsequent effects on radiation lead to an important chemistry-climate feedback~\cite{wang2025ozone}. For example, ozone changes have been demonstrated to play a key role in modulating projected climate responses -- from stratospheric cooling to surface warming~\cite{jonsson2004doubled,chiodo2019jc,Dietmuller2014,Muthers2014gmd}. By altering meridional and vertical temperature gradients, ozone is also projected to affect changes in the stratospheric and tropospheric circulation~\cite{jonsson2004doubled,chiodo2019jc}, including the QBO~\cite{tian2006quasi,DallaSanta2021jgr}, jet stream positions and strengths~\cite{nowack2018jgr,chiodo2019jc,Haase2020acp}, ENSO and the Walker circulation~\cite{nowack2017grl,nowack2018jgr}, Northern Atlantic Oscillation (NAO)~\cite{kuroda2008role,Chiodo_Polvani2019}, and the polar vortices~\cite{oehrlein2020effect}. Therefore, realistically representing the two-way coupling of ozone and climate is critical for reliable climate simulations.

However, the treatment of ozone in climate models is often internally inconsistent with the climate model state and the climate change scenario. The main reason is that interactive atmospheric chemistry schemes are typically too computationally expensive, slowing down global climate change simulations by more than a factor of two~\cite{horowitz2020gfdl,Archibald2020}. An explicit representation of ozone requires the model to calculate chemical tendencies based on complicated mathematical formulations and transport processes. As a result, only 21 out of the 60 Coupled Model Intercomparison Project Phase 6 (CMIP6) models include interactive ozone in DECK (including AMIP, pre-industrial, abrupt-4x${\mathrm{CO}_2}$, and 1\%/year ${\mathrm{CO}_2}$ increase) experiments, despite these being among the highest-priority simulations in the CMIP framework~\cite{IPCC2021}. Models without interactive ozone chemistry typically prescribe ozone (“non-interactive ozone”) at pre-industrial or present-day levels, not including the ozone response to GHG forcings~\cite{Cionni2011,keeble2021evaluating,DallaSanta2021jgr}. Unlike for historical or future scenario experiments, there are also no recommended datasets for prescribing ozone concentrations in the climate sensitivity experiments within DECK, including the abrupt-4x${\mathrm{CO}_2}$ and 1\%/year ${\mathrm{CO}_2}$ increase experiments~\cite{Input4MIP2016}. Moreover, among the aforementioned 21 models with interactive ozone scheme, five use the simplified linearized ozone photochemistry scheme (Linoz~\cite{mclinden2000stratospheric}), which is a common ozone parameterization scheme due to its low computational cost~\cite{mclinden2000stratospheric}. However, because of its assumptions, in particular on linearization of temperature and net chemical production~\cite{meraner2020useful}, it has several important limitations. These include significant bias in column ozone representation~\cite{meraner2020useful}, underestimation on QBO- and extratropical quasi-stationary planetary waves-related ozone variability~\cite{meraner2020useful}, non-applicability in the troposphere~\cite{meraner2020useful}, and discontinuity between polar and lower latitude regions~\cite{mclinden2000stratospheric}, etc.

For both efficiency and accuracy, we propose using machine learning (ML) to represent ozone interactively in climate models. \citeA{Nowack2018erl} introduced an ML-based ozone parameterization that generates three-dimensional daily ozone fields for climate sensitivity simulations, using temperature as the sole input. Despite its simplicity, the scheme achieves accurate and robust offline predictions, owing to the strong connections between ozone and atmospheric temperatures. \citeA{nowack2019machine} further demonstrated the possibility to transfer the ML parameterization across climate models, subject to straightforward data transformations. This study implements this idea for the first time interactively, both in the UK Earth System Model (UKESM) and the ICOsahedral Nonhydrostatic model (ICON) climate modeling frameworks. Such an implementation introduces major challenges, in particular that the ML scheme must be able to realistically represent ozone variability and trends on long, decadal timescales when two-way coupled to the physical model state. We will demonstrate that the ML scheme allows for an interactive representation of ozone for climate sensitivity experiments, with negligible cost compared to full-chemistry schemes such as the United Kingdom Chemistry and Aerosols (UKCA) Stratosphere-Troposphere scheme~\cite{Archibald2020}, by avoiding both the comprehensive computations on chemistry and transport processes.

Previous studies have shown successful applications of ML in predicting scenario-dependent ozone changes in the troposphere and stratosphere, but none have been implemented into a climate model, e.g., Bayesian Neural Network (NN) for historical ozone simulations~\cite{sengupta2020ensembling}, linear and non-linear ML techniques for future ozone projections under emission scenarios~\cite{keeble2021evaluating}. \citeA{Mohn2023eds} proposed a NN-based scheme for a daily ozone parameterization in atmospheric models, but their approach relied on certain catalytic chemical compounds as input, which cannot be obtained without running a full chemistry module or need to be specified separately as climatologies or from look-up tables. In a broader sense, ML has been gaining popularity as an optimized approach to parameterize other components in climate models in recent years, especially sub-grid processes including gravity waves~\cite{pahlavan2024explainable} and cloud cover~\cite{grundner2022deep}, and highly time-consuming components such as aerosols~\cite{kumar2024mieai}. Overall, the application of ML approaches to stratospheric ozone research is still in its infancy. To the best of our knowledge, there has been no successful online implementation of an ML scheme for parameterizing global ozone distributions before.

In this article, we assess the effectiveness of a ML–based ozone parameterization (mloz) in reproducing (a) the temporal variability and climatological spatial distribution of ozone, and (b) the ozone response and feedback to a quadrupling of atmospheric ${\mathrm{CO}_2}$. The ML scheme is first implemented in UKESM and evaluated against the UKESM full chemistry module -- UKCA~\cite{Archibald2020} -- regarding its accuracy and efficiency. Then its transferability is assessed by applying the ML scheme trained with full chemistry data from UKESM to the ICON model.

\section{Methods}
\subsection{The mloz scheme}
\begin{figure}[ht]
    \centering
    \includegraphics[width=0.5\linewidth]{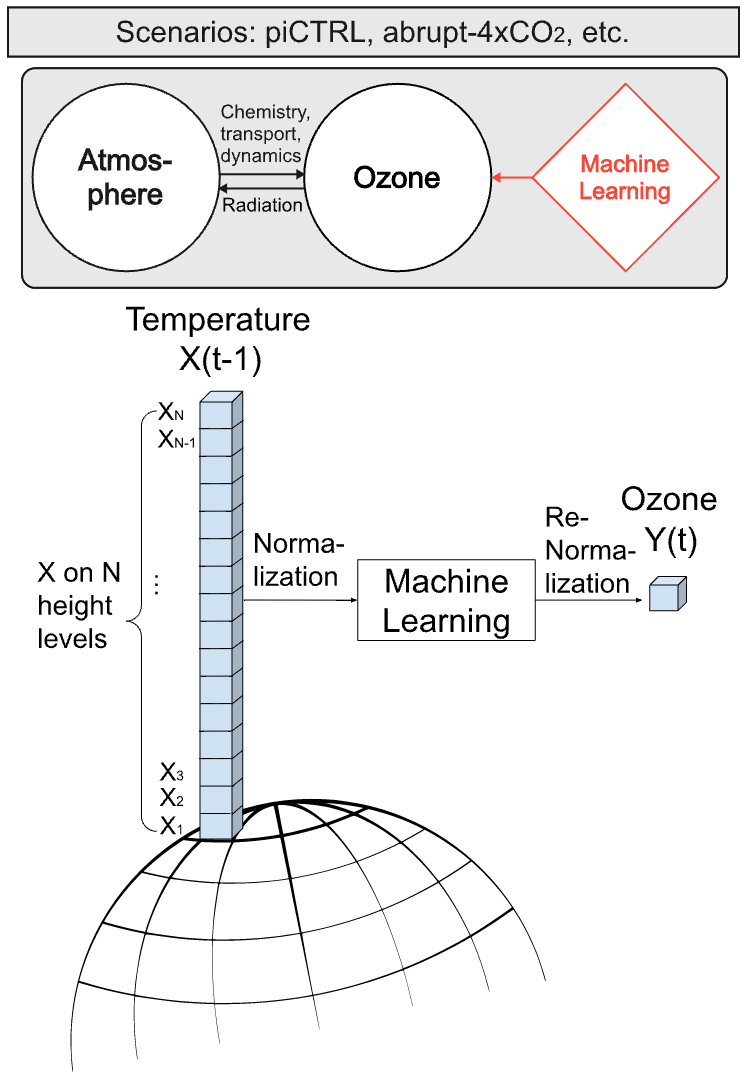}
    \caption{A schematic representation of the mloz parameterization. We use a machine learning (ML) approach to represent ozone and the associated two way climate-ozone coupling in climate models. For each ozone target grid point, we train a ML function independently for the mapping between single column temperature at timestep $t-1$ and single point ozone at timestep $t$. Then the offline-trained mapping is applied for online testing in climate models. N indicates the number of vertical levels of the single column temperature input. Here N=76 for the mloz in UKESM; N=71 for the mloz in ICON. Comprehensive offline comparisons showed that linear ridge regression outperforms other conventional nonlinear methods on this task. The use of column-only temperature information facilitates the parallelization on high-performance computing systems.}
    \label{fig:fig1}
\end{figure}

Figure~\ref{fig:fig1} illustrates the mloz scheme, which uses a ML approach to parameterize ozone up to 50~km altitude. Above this altitude, a climatology from the UKESM full-chemistry simulation is applied to stabilize the upper atmospheric climate condition and to prevent instabilities caused by input-output error accumulation. The mloz represents ozone interactively, including the ozone-circulation coupling and ozone's radiative feedback. In essence, mloz uses ML to map meteorological variables from the previous timestep to the ozone volume mixing ratio at the current timestep, where we here choose daily mean time resolution. Ideally, these input variables should be independent of the chemistry module, allowing ozone prediction without solving the underlying system of coupled partial differential chemical rate equations~\cite{pyle1980calculation}. In this study, we only use temperature as input, as it is readily available in general circulation models without atmospheric chemistry components. We adopt a single column input scheme in which we consider temperatures at all vertical levels within the same column as the target ozone grid point. Temperature reflects key processes influencing ozone in standard climate sensitivity simulations -- including production, depletion, and transport -- making it a strong predictor for ozone trends and variability, especially in the stratosphere. We offline tested additional predictor variables (e.g., humidity and pressure), which helped with predictive skill to a degree (Figure~S1 in Supporting Information S1), but reduced online stability probably due to error accumulation stemming from the dynamic coupling between predictors and predictands. We thus favor the temperature-only implementation here, also due to its advantageous in simplicity and smaller RAM requirements. Additionally, we find that expanding the spatial extent of input features -- such as using box, zonal or global field, can significantly enhance the offline skill (Figure~S2 in Supporting Information S1), but would make parallelization on massively parallel high-performance computing systems more difficult, sacrificing some computational speed-up and portability, so that we stay with the column-wise input scheme in this paper. The ozone predictions are constrained as positive values by replacing all negative values by 0, albeit negative predictions barely appear.

Various ML methods, ranging from linear ML methods including ridge and lasso regression, to non-linear methods including Random Forests, Long Short-Term Memory (LSTM) and a variety of other types of Neural Networks (NNs), have been evaluated and also previously reported~\cite{Nowack2018erl,nowack2019machine}. We also ran several hackathon events with more than 100~participants to consolidate the evaluations. Despite their higher model capacities, non-linear approaches did not outperform ridge regression offline. Only in ensemble models with ridge regression as baseline model could superior performance be achieved, but such approaches would disproportionally affect runtime performance given relatively small gains in offline accuracy. For example, our offline evaluation shows that, with the same input and output features, ridge regression performs as good as feedforward NNs in most areas of the atmosphere, and slightly outperform artificial NNs in the tropical upper stratosphere (Figure~S3 in Supporting Information S1). The superior performance of ridge regression can be explained by the strong linear relationship between temperature and ozone in most regions of the atmosphere~\cite{Moreira2016acp,Randel2021acp,Hocke2023atmo}. Given its intrinsic interpretability, computational efficiency, and ease of implementation, the ridge regression approach is chosen here for the parameterization.

We use ridge regression to map the temperature over a single column to the ozone at a single target grid point (Figure~\ref{fig:fig1}). Ridge regression is a linear regression augmented by L2-regularization, which penalizes large coefficient values and has the advantage of addressing overfitting compared to a pure linear regression model. The strength of the regularization term is controlled by the hyperparameter \(\alpha\). For each target grid point over all latitudes, longitudes and height levels, the ridge coefficients (\(c_1\) to \(c_N\)) minimize a penalized residual sum of squares, :

\begin{equation}
    J^{ridge} = \arg\min_{c} \left\{ \sum_{i=1}^{n} \left( y_i - \sum_{j=1}^{N} x_{ij} \, c_j \right)^2 + \alpha \sum_{j=1}^{N} c_j^2 \right\}
\end{equation}

Here, \(x_{ij}\) is the normalized temperature value at the \(j^{th}\) height level for the \(i^{th}\) sample, \(y_{i}\) is the normalized ozone prediction for the same sample, and N is the number of vertical levels up to 50~km in a climate model. The parameters are validated with three-fold cross-validation. The hyperparameter \(\alpha\) is optimized according to the averaged generalization performance on validation sets. Final coefficients are refitted on the entire dataset subject to the optimal cross-validated $\alpha$ to make full use of all available data. In the end, these result in a 4-dimensional array of ridge coefficients, where each latitude-longitude-height grid point is associated with a coefficient series (\(c_1\) to \(c_N\)).

The implementation and evaluation of mloz is divided into two stages:

\begin{enumerate}[label=(\roman*)]
    \item Offline Training: For each model grid point at a given latitude, longitude and vertical level, a set of ridge regression coefficients \(c_j,j = 1,\ldotp \ldotp \ldotp ,N\) and hyperparameter \(\alpha\) are trained on 40 years of UKESM full-chemistry simulations from Experiment A and Experiment B, separately. During this process, scaling parameters (mean and standard deviation) for temperature and ozone are also saved for use in the online implementation.
    \item Online Testing in ICON and UKESM: A daily prediction timestep is used, where the ozone volume mixing ratio is predicted using the previous day’s mean temperature profile. Thus, the mloz calculation is performed once per model day. Pre-trained regression coefficients are read from a standard NetCDF4 file and loaded at the start of the model run. Prior to the regression computation, daily mean temperatures from model level 1 to N are standardized to unit variance. The (standardized) ozone value at each grid cell on the current day is then computed via matrix multiplication between the (standardized) temperature vector on previous day and the corresponding ridge coefficients along the vertical column: \(\widehat{y} = \sum_{j=1}^{N} x_{j} \, c_j\). Afterward, the predicted ozone value is re-scaled back to mixing ratio unit. This computation is parallelized across grid cell blocks to ensure efficient ozone prediction throughout the global domain.
\end{enumerate}

This simple regression-based scheme makes mloz particularly computationally efficient, easily runs in most parallel model environments, and is straightforward to implement across different climate models. For the transfer into ICON, a recalibration step is applied to the temperature input during the scaling process, as described in Section 2.3.

\subsection{Climate models setup}
In the online testing stage, the trained function is first applied interactively in UKESM1~\cite{Sellar2019} to produce daily ozone simulations. These parameterized simulations are then assessed by comparing their statistics against those from the full-chemistry UKESM runs. The UKESM includes coupled components for the atmosphere, ocean, sea ice, land surface, and atmospheric chemistry. The experiments with UKESM are configured as Coupled Model Intercomparison Project (CMIP)-like simulations. Simulations are run for 50 years to allow the model to level off towards an approximate equilibrium state in the last 20 years under 4x${\mathrm{CO}_2}$ forcing (Table \ref{tab:table1}). 

\begin{table}[h]
\caption{Experimental set-up for the simulations.}
\label{tab:table1}
\begin{tabular}{llllll}
\toprule
Model & \begin{tabular}[c]{@{}l@{}}Exper-\\ iment\end{tabular} & Forcing & Ozone chemistry & Ocean & \begin{tabular}[c]{@{}l@{}}Simulation\\ years\end{tabular} \\ 
\midrule
\multirow{4}{*}{UKESM} & A & \multirow{2}{*}{piCTRL} & full chemistry & \multirow{4}{*}{\begin{tabular}[c]{@{}l@{}}interactive \\ ocean\end{tabular}} & \multirow{4}{*}{50} \\
 & A1 &  & mloz &  &  \\\cmidrule{2-4}
 & B & \multirow{2}{*}{\begin{tabular}[c]{@{}l@{}}abrupt- \\ 4x${\mathrm{CO}_2}$\end{tabular}} & full chemistry &  &  \\
 & B1 &  & mloz &  &  \\\cmidrule{1-6}
\multirow{7}{*}{ICON} & A2 & \multirow{3}{*}{piCTRL} & mloz & \multirow{3}{*}{\begin{tabular}[c]{@{}l@{}}prescribed SSTs \\ climatology (last \\ 20 years of A)\end{tabular}} & \multirow{7}{*}{30} \\
 & A3 &  & Linoz &  &  \\
 & A4 &  & prescribed piCTRL ozone &  &  \\\cmidrule{2-5}
 & B2 & \multirow{4}{*}{\begin{tabular}[c]{@{}l@{}}abrupt- \\ 4x${\mathrm{CO}_2}$\end{tabular}} & mloz & \multirow{4}{*}{\begin{tabular}[c]{@{}l@{}}prescribed SSTs \\ climatology (last \\ 20 years of B)\end{tabular}} &  \\
 & B3 &  & Linoz &  &  \\
 & B4 &  & prescribed piCTRL ozone &  &  \\
 & B5 &  & prescribed 4x${\mathrm{CO}_2}$ ozone &  &  \\ 
\bottomrule
\end{tabular}
\end{table}

UKESM employs the United Kingdom Chemistry and Aerosols (UKCA~\cite{Archibald2020}) module, which allows for the generation of training data for mloz from its comprehensive stratospheric-tropospheric chemistry scheme~\cite{Archibald2020}. It simulates 306 reactions among 87 species, with lightning ${\mathrm{NO}_\mathrm{x}}$ emissions parameterized following~\cite{price1992simple} and photolysis rates computed using the FastJX scheme~\cite{Telford2013}. The radiative effects of ozone are interactively coupled to the model dynamics, enabling feedbacks between atmospheric composition and climate.

The atmospheric component of UKESM is the Met Office’s Unified Model with a horizontal resolution of 1.875° longitude by 1.25° latitude and 85 vertical levels extending to 85 km. Subgrid-scale features such as clouds and gravity waves are parameterized, and the version used here includes a self-contained QBO. The ocean component is based on European Modeling of the Ocean model version 3.6~\cite{madec2017nemo}, using a tripolar grid with a 2° longitude resolution and enhanced latitudinal resolution (up to 0.5°) in the tropics.

Subsequently, we apply the function in another climate model -- ICON -- to verify its transferability, which might open up pathways to wide applicability across climate models. The ICON experiments are configured as Atmospheric Model Intercomparison Project (AMIP)-like simulations~\cite{Niemeier2023grl}, using the configuration of ICON 2024.07 release~\cite{Wolfgang2025}, with the atmospheric component ICON-NWP~\cite{Zängl2015} coupled with the land component JSBACH 4. The ICON simulations are implemented with prescribed monthly climatological sea surface temperature and sea ice concentration derived from the last 20 years of UKESM full chemistry simulations (Table~\ref{tab:table1}). 

In this study, we compare mloz with Linoz and assess ozone responses to abrupt-4x${\mathrm{CO}_2}$ forcing by evaluating against simulations with prescribed ozone (see Table~\ref{tab:table1}). ICON is run at an icosahedral resolution of R2B5, corresponding to approximately 80~km horizontal resolution. Vertically, the model includes 130 levels, extending from the surface up to around 75 km. Layer thickness increases with altitude, ranging from about 10 m near the surface to $\sim$500m at 35km and approximately 1 km at the stratopause. This model setup is adapted for a good representation of stratospheric dynamics and transport~\cite{Niemeier2023grl}.

\subsection{Re-grid and recalibration for the transfer}

Transferring mloz from UKESM to ICON requires regridding and recalibration due to differences in vertical coordinate systems and climatological states. Before the ML calculation, temperature inputs are interpolated onto the UKESM height levels primarily using cubic spline interpolation. Afterwards, the ozone fields are bilinearly interpolated back onto ICON’s vertical grid. These interpolation weights are computed once at the beginning of the model run to minimize computational overhead. To avoid extrapolation errors, particularly in regions where ICON’s Earth surface definition lies under UKESM’s, the lowest six ICON levels (below 270m) are filled with climatological ozone values from UKESM full chemistry. Additionally, to mitigate the inevitable differences in temperature and ozone fields between two models, a simple recalibration is applied during the standardization step in ICON, as described in Equations (\ref{eq2})~\cite{nowack2019machine}.

\begin{eqnarray}
\label{eq2}
   X^*_{ICON} = (X_{ICON} - \overline{X}_{ICON}) / X^{std}_{ICON} \nonumber \\
   Y_{ICON} = \hat{Y}^*_{ICON} \times Y^{std}_{UKESM} + \overline{Y}_{UKESM}
\end{eqnarray}

Here X, Y represent temperature and ozone, respectively. \(\overline{X}_{ICON}\) and \(X^{std}_{ICON}\) are the climatology and standard deviation of temperature from the ICON run prescribed by SSTs and ozone from UKESM. By substracting \(\overline{X}_{ICON}\) from \(X_{ICON}\), the ICON temperature is recalibrated into an approximately zero mean (for piCTRL runs). The \(Y^{std}_{UKESM}\) and \(\overline{Y}_{UKESM}\) are obtained from UKESM training datasets. Interestingly, we find that temperature scalings used for the transfer recalibration from piCTRL are applicable to the 4x${\mathrm{CO}_2}$ run.

\section{Results}
\subsection{Online performance in multi-decadal pre-industrial simulations}

An important baseline performance test for any parameterization is its capability to produce reasonable internal variability on daily to decadal timescales. For this purpose, the benchmark experiments in climate modeling is pre-industrial control (piCTRL) simulations, i.e. runs without anthropogenic influences and an assumed constant ${\mathrm{CO}_2}$ concentration (ca. 280~ppmv). A summary of the model experiments in this study is shown in Table~\ref{tab:table1}. We run UKESM with the mloz parameterization for CMIP-like simulations and compare the results with those obtained using the full UKCA chemistry scheme, assessing how well mloz captures variability and grid point-wise distributions of ozone.

\begin{figure}
    \centering
    \includegraphics[width=0.9\linewidth]{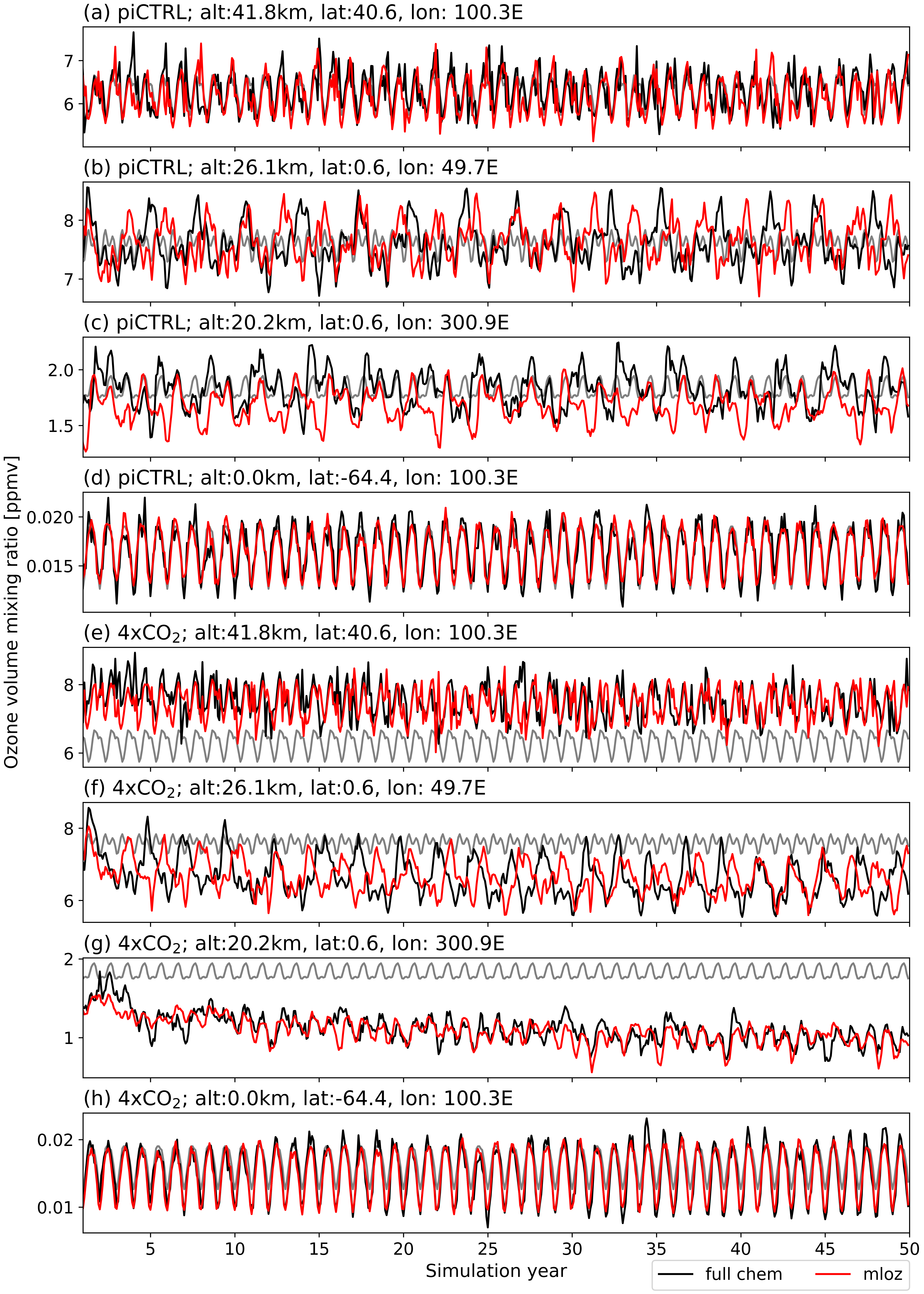}
    \caption{UKESM time series of mloz predictions on randomly sampled, but representatively selected grid points. Red lines show mloz-predicted ozone, black lines represent full-chemistry ozone from UKESM, and grey lines indicate the fixed preindustrial climatology from UKESM’s full-chemistry simulation (Experiment A). For clarity, all time series shown are averaged to monthly means. Black/red lines in (a)-(d) are from Experiment A/A1 piCTRL simulations and those in (e)-(h) are from Experiment B/B1 4x${\mathrm{CO}_2}$ simulations. Ozone values are provided in ppmv. Note that the ozone time series on selected points in (b), (c), (f), and (g) depict QBO-related variability and mloz successfully captures the quasi-biennial variability in ozone, though not in-phase due to the nature of a free running climate model. The grey lines illustrate the ozone scheme that climate models without a chemistry module commonly adopt for climate sensitivity simulations -- prescribed ozone at a fixed preindustrial annual climatology. The ozone output in the 1-year spin-up period is not shown in the plot.}
    \label{fig:fig2}
\end{figure}

Overall, mloz produces stable ozone predictions over 50 years and captures all major aspects of variability, as illustrated by four randomly selected grid-point-wise time series in Figures~\ref{fig:fig2}(a)-(d). Stable ozone output indicates that biases do not accumulate throughout the model run; otherwise, the ozone volume mixing ratios would deviate from climatology and induce increasing biases in the temperature background through feedback processes. mloz can easily reproduce seasonal variability across different atmospheric regions (Figures~\ref{fig:fig2}a-d) and captures Quasi-Biennial Oscillation (QBO)-related quasi-biennial variability in the low-mid stratosphere (Figures~\ref{fig:fig2}b,c). The QBO, alternating easterlies and westerlies in the tropical stratosphere, dominates the interannual variability of tropical stratospheric ozone~\cite{tian2006quasi}. In contrast, the prescribed ozone climatology (grey lines in Figures~\ref{fig:fig2}) -- commonly used in climate sensitivity simulations by models without interactive chemistry -- cannot represent QBO-related variability and ozone changes in response to increasing ${\mathrm{CO}_2}$ forcing. To be noticed, a perfect match for internal variabilities between mloz and the full chemistry module cannot be expected in a free-running simulation, as the memory of initial conditions typically dissipates after a few years~\cite{ma2022pacific}. Nevertheless, their statistical characteristics -- such as climatologies, probability density functions (PDFs), and standard deviations -- can still be meaningfully compared on sufficiently large sample sizes.

\begin{figure}[ht]
    \centering
    \includegraphics[width=0.6\linewidth]{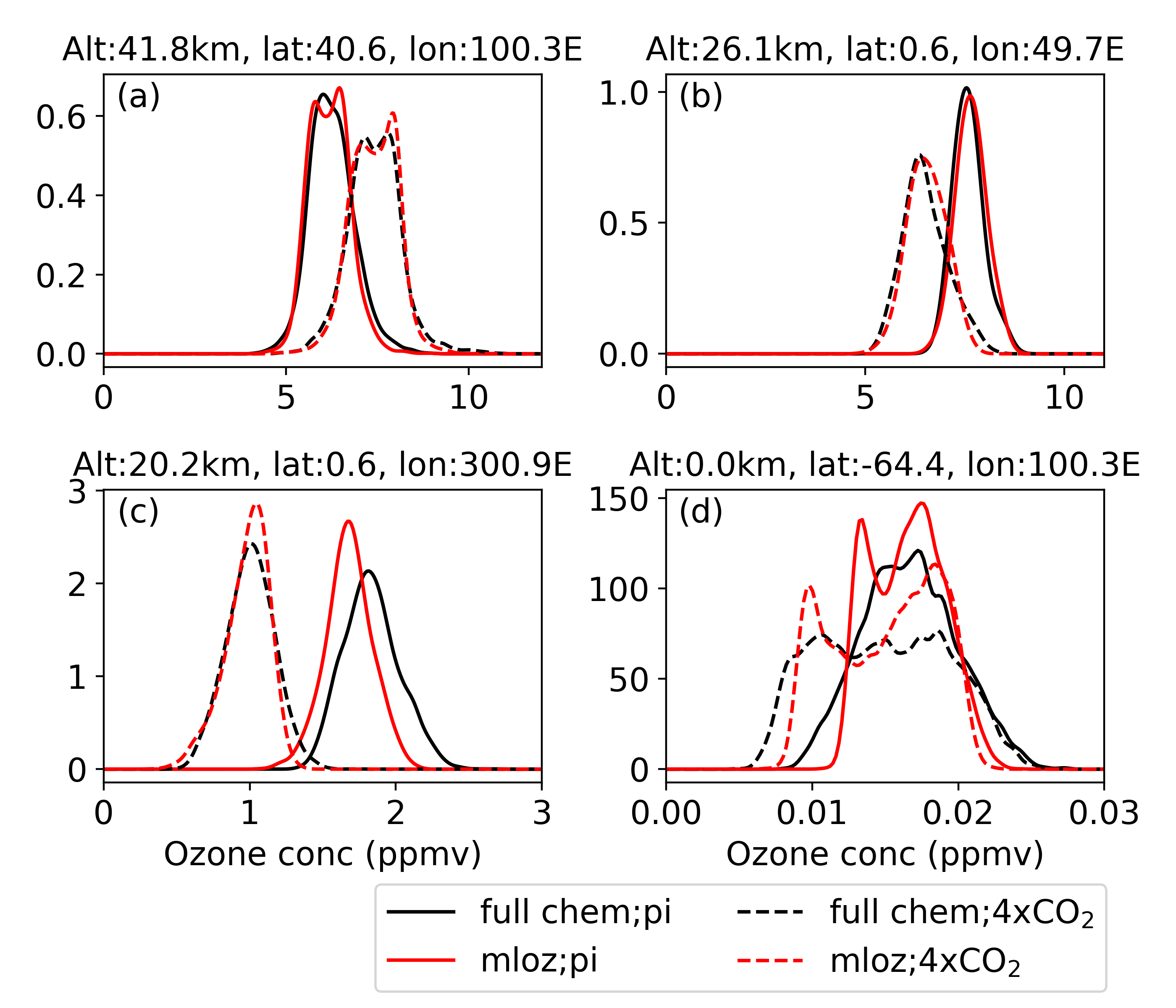}
    \caption{Probability Density Function (PDF) distributions of mloz ozone predictions from UKESM on the same grid points as in Figure~\ref{fig:fig2}. Red and black lines correspond to mloz and full chemistry simulations from UKESM, respectively. Solid and dashed lines denote the piCTRL and 4x${\mathrm{CO}_2}$ experiment, respectively. The PDFs are derived using Gaussian kernel density estimation, with the bandwidth set to 2\% of the average piCTRL UKESM full-chemistry ozone value at each grid point.}
    \label{fig:fig3}
\end{figure}

As shown in the plot of PDF distributions of ozone from mloz compared with full chemistry on these representative points (Figure~\ref{fig:fig3}), mloz generally represents the PDF distributions of ozone well in the stratosphere. This good match in PDF distributions between mloz and full chemistry scheme not only exist on these grid points, but also across the stratosphere, as can be seen from the distributions for ozone predictions across the lower and middle stratosphere (Figure~S4 in Supporting Information S1). As expected, the tropospheric ozone variability is harder to capture (Figure~\ref{fig:fig3}d), given limited predictability one day in advance and given the more complex ozone chemistry and a greater number of influencing factors, which can be uncorrelated with the temperature state. For example, lightning, which is difficult to predict one day in advance, triggers ozone production by initiating ${\mathrm{NO}_\mathrm{x}}$ emissions in the troposphere~\cite{verma2021role}. In contrast, the stratosphere is less influenced by high-frequency weather systems and is more affected by the highly predictable insolation and a more slowly moving circulation (BDC, QBO). However, ozone variability (rather than climate trends) in the lower troposphere is less important compared to the stratosphere, due to its significantly lower concentrations and its relatively minor influence on the tropospheric circulation~\cite{monks2015tropospheric,DallaSanta2021jgr,chiodo2019jc}. Overall, mloz represents ozone amplitudes well in lower latitudes, while underestimates ozone variability in the polar stratospheric regions to some degree (Figure~S5 in Supporting Information S1).

\begin{figure}[ht]
    \centering
    \includegraphics[width=1\linewidth]{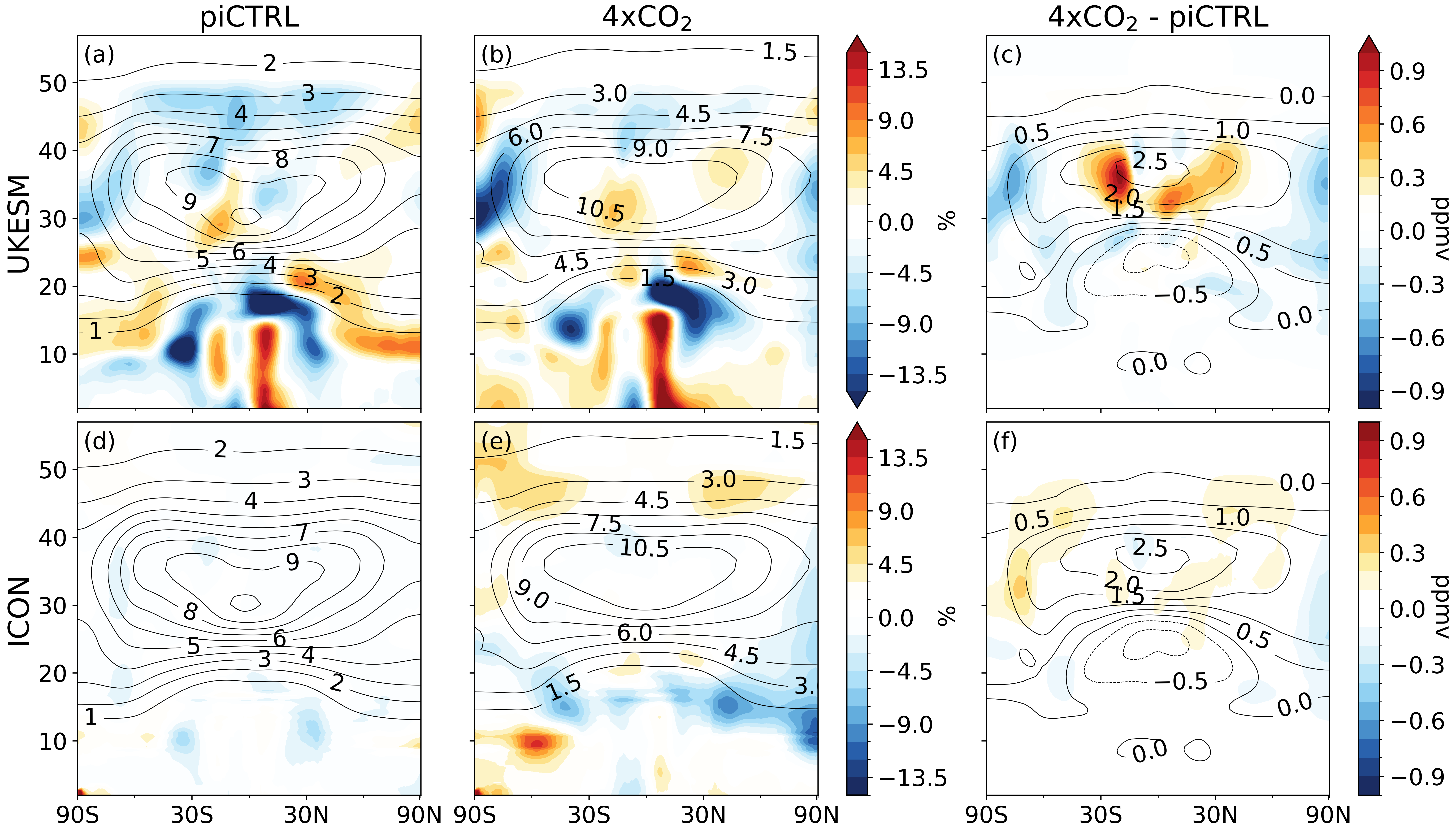}
    \caption{Long-term bias in ozone from mloz. Colors in the first two columns are the percentage differences in ozone climatology between mloz and the full chemistry simulation, specifically, Experiment A1 versus Experiment A in (a), Experiment B1 versus Experiment B in (b), Experiment A2 versus Experiment A in (d), and Experiment B2 versus Experiment B in (e). Colors in the third column are the volume mixing ratio differences between the ozone response with mloz and the ozone response with full chemistry to 4x${\mathrm{CO}_2}$. Contour lines indicate the UKESM full chemistry ozone, with lines in the first, second, and third column corresponding to Experiment A, Experiment B , Experiment B minus Experiment A, respectively. The UKESM ozone climatologies are averages over the last 20 years of the interactive-ocean simulations, and the ICON climatologies are averages over the 30 years of prescribed SSTs simulations. All fields presented are zonal means.}
    \label{fig:fig4}
\end{figure}

In addition to accurately capturing grid-point-wise variability, the mloz scheme effectively reproduces spatial distributions of ozone, including both zonal and horizontal patterns. Above all, any ozone parameterization should be able to reproduce the long-term climatology of an interactive chemistry module with high fidelity. Figure~\ref{fig:fig4}(a) shows the percentage differences in ozone volumn mixing ratios between mloz and the full chemistry module in UKESM piCTRL simulations, averaged over the final 20 years of each simulation. The bias in the ozone climatology produced by mloz is within 10\% throughout the stratosphere, which is much smaller than the spread in ozone climatologies found across, e.g., CMIP6 models~\cite{keeble2021evaluating}. A significant positive bias occurs between 10°S and 20°S in the middle stratosphere, indicating a slight asymmetry with the peak tilting towards the Southern Hemisphere. Notably, this bias is absent in the offline ozone prediction by mloz (Figure~S6 in Supporting Information S1). This highlights a broader characteristic of ML-based parameterizations, consistent with previous studies: their online performance cannot be reliably inferred from offline metrics alone~\cite{Reimers2025, mansfield2024uncertainty}. In the upper troposphere and lower stratosphere (UTLS), the mloz generally exhibits larger percentage deviations in ozone climatology  (Figure~\ref{fig:fig4}a). The percentage bias in the troposphere is also notably larger than in the stratosphere, reflecting reduced model skill in this region. However, this is partly due to the lower absolute ozone concentrations in the troposphere (Figure~\ref{fig:fig3}d; Figure~\ref{fig:fig2}d), where even small absolute deviations can result in relatively large percentage differences (Figure~\ref{fig:fig4}a). Finally, we demonstrate that mloz also faithfully reproduces the horizontal pattern of column ozone -- the amount of ozone contained in a vertical column of the atmosphere within a unit area. This is an important quantity of the atmosphere in general and resembles conservation quantities such as energy, water, and momentum used for the evaluation of Earth System Models (ESMs). It is also of central importance for ultraviolet exposure at Earth's surface and thus a central quantity in the chemistry-climate modeling community. The long-term climatological bias in column ozone from mloz is within 7.5\% globally (Figure~S7 in Supporting Information S1), demonstrating superior performance compared to other simplified linearized ozone schemes which exhibit biases up to 10\% (e.g., \cite{meraner2020useful}). This performance is also favorable when compared to the significant uncertainty among CMIP6 models, which show a spread exceeding 15\%~\cite{keeble2021evaluating}.

\subsection{Transferability from UKESM to ICON}

Given the lack of interactive chemistry schemes in the majority of global climate models~\cite{IPCC2021}, we identify the potential to build more widely applicable ozone parameterizations as an important challenge. In particular, any such parameterization should run stably when applied across different climate models, and be easily adaptable to each model's specific climate state. Here, we explore the possibility of transferring the mloz parameterization trained on a climate-chemistry model (UKESM) to a model without a full chemistry module (here we take the ICON without the full chemistry component as an example), starting with an evaluation on a piCTRL simulation. To eliminate the inherent differences in the temperature climatologies and variability between UKESM and ICON, we apply a recalibration to the temperature input for mloz in ICON by standardizing the ICON temperature input with scalings from ICON (see Section 2.3). We show that, after recalibration, the ridge functions trained on UKESM data can be successfully applied to ICON. Similar to UKESM, ICON does not show any instability in ozone predictions over 30 years (Figure~S8 in Supporting Information S1), and emulates an ozone climatology very close to UKESM full chemistry. Additionally, its long-term errors in both zonal and horizontal distributions relative to UKESM full chemistry ozone are minimal (Figure~\ref{fig:fig4}d, Figure~S7c in Supporting Information S1).

Somewhat surprisingly, percentage errors in the ozone climatology in ICON with mloz compare even more favorably to the ground truth UKESM data. A possible explanation is that errors in the ozone predictions are better buffered by the ICON temperature field than by the UKESM temperature field. As shown in Figure~\ref{fig:fig4}d, the bias in ozone climatology from the transferred mloz is less than 2.5\% in the stratosphere, and it also better reproduces tropical ozone. The bias in the troposphere is slightly higher but remains within 6\%. The column ozone simulation (Figure~S7c in Supporting Information S1) is also better than UKESM mloz, with a bias of less than 4.5\%.

\subsection{Representation of ozone response to abrupt-4x${\mathrm{CO}_2}$}

Next, we evaluate the performance of mloz in capturing the ozone response to climate change. With increased ${\mathrm{CO}_2}$, ozone should increase in the mid-upper stratosphere and decrease in the tropical lower stratosphere (Figure~S9i in Supporting Information S1)~\cite{keeble2021evaluating,keeble2017diagnosing,eyring2013long}. These changes obviously cannot be presented in simulations with fixed ozone climatologies (i.e., piCTRL ozone climatologies frequently used in abrupt-4x${\mathrm{CO}_2}$ CMIP runs), with important implications for the overall climate response~\cite{jonsson2004doubled,nowack2015large}. The question is to what extent can mloz represent such responses in UKESM and ICON. Here, we train and validate a new set of ridge coefficients with UKESM abrupt-4x${\mathrm{CO}_2}$ full chemistry simulations over 40~years following the abrupt ${\mathrm{CO}_2}$ increase. Note that different from piCTRL ozone, there is a centennial-millennial trend in abrupt-4x${\mathrm{CO}_2}$ ozone as a response to the ${\mathrm{CO}_2}$ forcing~\cite{Nowack2018erl}. Therefore, the mloz needs to be capable of capturing the additional climate change trend in order to correctly represent the ozone climatologies over the test period (as shown in Figure~\ref{fig:fig4}b,e). The linear ML algorithm used here is good at capturing the linear climate change trend, which could also make it a better extrapolation tool than other non-linear algorithms~\cite{Nowack2018erl,Nowack2021AMT,nowack2025acp}.

The mloz parameterization demonstrates comparable skill in representing ozone under the abrupt-4x${\mathrm{CO}_2}$ scenario as in the piCTRL simulations (Figure~\ref{fig:fig3}, Figures~\ref{fig:fig4}b,e), despite the added challenge of capturing additional climate forcing-driven trends and the trends beyond the training data time horizon. Notably, mloz successfully reproduces these trends, as illustrated at several representative grid points (Figures~\ref{fig:fig2}f,g). This is also reflected in the good match in statistical distributions of abrupt-4x${\mathrm{CO}_2}$ ozone from mloz and full chemistry scheme in different regions of the atmosphere, whose averages are distinct from that of piCTRL ozone (Figure~\ref{fig:fig3}a-c, Figure~S4 in Supporting Information S1). The long-term bias for 4x${\mathrm{CO}_2}$ runs is constrained to within 10\% in most of the stratosphere in UKESM (Figure~\ref{fig:fig4}b), and 6\% in ICON (Figure~\ref{fig:fig4}e). However, similar to piCTRL, mloz tends to underestimate ozone variability above the polar regions under 4x${\mathrm{CO}_2}$ (Figure~S5b in Supporting Information S1). This leaves room for additional developments but is within typical uncertainties in CMIP6 multi-model experiments~\cite{keeble2021evaluating}. Consistent with the good representation of 4x${\mathrm{CO}_2}$ ozone, mloz can also represent ozone's response to 4x${\mathrm{CO}_2}$ (Figures~\ref{fig:fig4}c,f), which is the difference between 4x${\mathrm{CO}_2}$ and piCTRL ozone. The ozone response simulated by UKESM mloz exhibits a peak bias of 0.9 ppmv in the tropical middle stratosphere (Figure~\ref{fig:fig4}c). However, the ozone decrease in the tropical lower stratosphere as a response to 4x${\mathrm{CO}_2}$, which is particularly important for ozone feedback on climate sensitivity~\cite{nowack2015large}, is accurately represented by mloz (Figure~\ref{fig:fig4}c). This ozone decrease is primarily attributed to the acceleration of the BDC in response to 4x${\mathrm{CO}_2}$ and the expansion of the troposphere. For the troposphere, we cannot capture high frequency daily variability (Figures~\ref{fig:fig4}a,b), but we are able to faithfully represent forced ozone trends under changing climatic conditions (Figure~\ref{fig:fig4}c), which is important because of their effects on the global energy budget~\cite{IPCC2021Chapter6,WMO2022}. The transferred mloz in ICON well represents the response of ozone to 4x${\mathrm{CO}_2}$, with a deviation less than 0.35~ppmv across the whole atmosphere (Figure~\ref{fig:fig4}f).

\subsection{Representation of ozone feedback}

\begin{figure}
    \centering
    \includegraphics[width=0.5\linewidth]{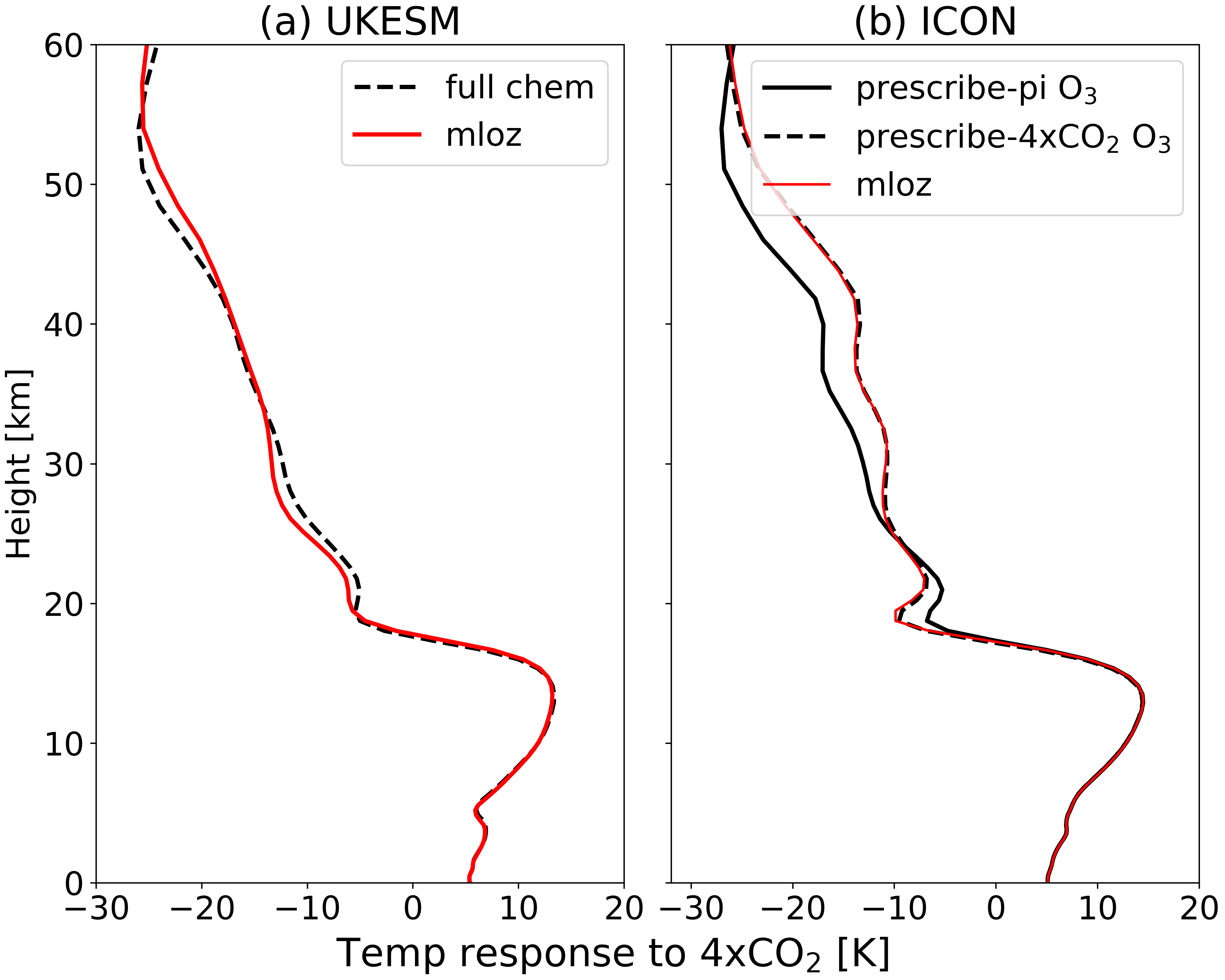}
    \caption{Profiles of equatorial temperature responses to 4x${\mathrm{CO}_2}$ with different ozone schemes. (a) Vertical profile of the equatorial zonal mean temperature response to 4x${\mathrm{CO}_2}$ from UKESM with mloz (red line) compared to with full chemistry (black dashed line). The temperature responses of mloz and full chemistry scheme are averaged over the last 20 years of simulations Experiment B1 minus Experiment A1 and Experiment B minus Experiment A, respectively. (b) Vertical profile of the equatorial zonal mean temperature response from ICON mloz (red line), with prescribed piCTRL ozone climatology from UKESM (black solid line), and with prescribed 4x${\mathrm{CO}_2}$ ozone climatology from UKESM (black dashed line). The discrepancy between the black solid and black dashed line indicates the bias in temperature simulation when ICON is prescribed with fixed piCTRL ozone climatology for the 4x${\mathrm{CO}_2}$ run. The temperature responses of mloz, prescribe-pi ozone, and prescribe-4x${\mathrm{CO}_2}$ ozone are averaged over the 30 years of simulations Experiment B2 minus Experiment A2, Experiment B4 minus Experiment A4, and Experiment B5 minus Experiment A4, respectively. In both (a) and (b), the closer the red and black dashed lines are, the more temperature modeling under ${\mathrm{CO}_2}$ forcing benefits from mloz interactive ozone.}
    \label{fig:fig5}
\end{figure}

Many climate modeling studies have demonstrated the importance of two-way interactions between ozone and climate change~\cite <i.e.>[]{jonsson2004doubled,Schröter2018gmd,nowack2015large,chiodo2019jc,Dietmuller2014,Muthers2014gmd}. Stratospheric ozone modifies the climate system's response to GHG increases, impacts temperatures from the stratosphere down to the surface, and influences stratospheric and tropospheric circulation responses to the GHG forcing, thus acting as a climate feedback~\cite{WMO2022}. As shown in Figure~\ref{fig:fig5}(b), the equatorial temperature response to 4x${\mathrm{CO}_2}$ differs in ICON if piCTRL (black solid line) or 4x${\mathrm{CO}_2}$ (black dashed line) ozone climatology from UKESM is prescribed. This is because ozone increases in the mid-upper stratosphere (Figure~S9i in Supporting Information S1) due to stratospheric cooling in response to 4x${\mathrm{CO}_2}$, which absorbs more solar ultraviolet radiation and mitigates local stratospheric cooling~\cite{jonsson2004doubled}. In contrast, ozone is expected to decrease under ${\mathrm{CO}_2}$ forcing in the tropical lower stratosphere (Figure~S9i in Supporting Information S1) due to an accelerated BDC and reduced ozone production caused by a thicker overhead column ozone~\cite{Meul2014acp}. This enhances stratospheric cooling in this region. Compared with fixed ozone, including interactive ozone results in 3-5~K less cooling in the upper stratosphere and enhanced cooling of 2-3~K in the lower stratosphere under 4x${\mathrm{CO}_2}$ (comparing solid and dashed black lines), consistent with previous studies~\cite <i.e.>[]{Dacie2019jc}. The interactive ozone configuration using the mloz scheme effectively captures this feedback in the ICON model (Figure~\ref{fig:fig5}b), although minor discrepancies remain in UKESM, particularly in the middle stratosphere (Figure~\ref{fig:fig5}a). This is likely due to more accurate representation of the ozone response to 4x${\mathrm{CO}_2}$ in ICON compared to UKESM (Figures~\ref{fig:fig4}c,f). The strong agreement between temperature response profiles from the mloz and prescribed-4x${\mathrm{CO}_2}$ ozone not only appears in equatorial climatology profile of temperature, but also in the time series (Figure~S10 in Supporting Information S1) and the background field (Figure~S11f in Supporting Information S1). Accordingly, the mloz scheme reproduces the ozone-induced feedbacks with high fidelity, maintaining biases below 1.6~K in the representation of temperature response throughout the troposphere and stratosphere, with a better performance in ICON than UKESM (Figures.~S11c,f in Supporting Information S1). Nonetheless, both models exhibit a common tendency to underestimate temperatures in the tropical upper troposphere and lower stratosphere (UTLS), and to overestimate temperatures in the tropical mid-to-upper stratosphere (Figures.~S11a,b,d,e in Supporting Information S1). Moreover, the close agreement in surface temperature response between mloz and full chemistry simulations (Figure~\ref{fig:fig5}a) highlights the scheme’s ability to capture the radiative impact of ozone. This is important because previous studies have demonstrated that ozone feedback can significantly moderate surface warming under 4x${\mathrm{CO}_2}$~\cite <i.e.>[]{Dietmuller2014,nowack2015large,nowack2018jgr}.

\subsection{Comparison to Linoz}

Given that Linoz is a standard ozone parameterization scheme in the current ICON model~\cite{Schröter2018gmd}, we here compare mloz with Linoz in ICON with respect to accuracy and efficiency. Note that since the Linoz table we use (see Methods) is not consistent with UKESM ozone chemistry, the UKESM full chemistry ozone cannot serve as benchmark. Nevertheless, qualitative comparisons still offer valuable insights into the representation of ozone patterns and responses. As expected, there are differences between the ozone climatologies produced by mloz (which is consistent with UKESM) and Linoz, in particular in terms of peak ozone concentrations and meridional gradients (Figures~S9d,e in Supporting Information S1). The ozone time series from mloz also fluctuates more closely around the UKESM full chemistry climatology, especially in troposphere and tropical UTLS region (Figures~S8c,d,g,h in Supporting Information S1). mloz closely reproduces the spatial pattern of ozone response relative to the UKESM full-chemistry simulation as shown in Figure~\ref{fig:fig4}(f). Linoz, on the other hand, tends to simulate a larger ozone increase in the troposphere in response to 4x${\mathrm{CO}_2}$ (Figure~S9h in Supporting Information S1), consistent with the findings from previous studies with Linoz~\cite{meraner2020useful}. This occurs because the temperature in the troposphere rises under 4x${\mathrm{CO}_2}$ conditions, causing the linearized scheme, which depends on local temperature, to incorrectly overestimate ozone. However, it should be noted that Linoz is not specifically designed for a realistic representation of tropospheric chemistry~\cite{meraner2020useful}, whereas there is no such limitation on a ML ozone scheme.

Table~\ref{tab:table2} summarizes the average wall-clock time for the mloz scheme compared to the UKCA chemistry module in UKESM and the Linoz scheme in ICON. In UKESM, the full chemistry module accounts for approximately 55\% of the total simulation time, with nearly half of that attributed to tracer transport processes~\cite{esenturk2018quasi}. This means that running interactive chemistry at least doubles the UKESM model runtime. In contrast, the mloz scheme requires only 1.75\% of the same simulation time, making it approximately 31 times faster than UKCA. This substantial gain in efficiency stems from bypassing the computationally intensive differential equations and transport calculations required for interactive chemistry. Instead, mloz uses only temperature as input and predicts ozone via matrix multiplication with pre-trained ridge regression coefficients. In ICON, the mloz scheme takes an average of 15.78 minutes for a 1.5-year AMIP-like simulation, accounting for just 2.42\% of the total runtime. Compared with Linoz, mloz is still 2.75 times faster. A key difference is that mloz does not just model chemical tendencies but also replaces transport costs for chemical tracers.

\begin{table}[h]
\caption{Average wall-clock time across all processors for different chemistry schemes.}
\label{tab:table2}
    \centering
    \begin{tabular}{c|cc|cc}
    \toprule
Model & \multicolumn{2}{c|}{UKESM} & \multicolumn{2}{c}{ICON} \\
Chemistry scheme & mloz    & chemistry module   & mloz       & Linoz       \\
    \midrule
Cores            & \multicolumn{2}{c|}{720}   & \multicolumn{2}{c}{912}  \\
Length         & \multicolumn{2}{c|}{3 months}   & \multicolumn{2}{c}{1.5 years}  \\
Time cost        &  1.75\% & 54.45\%~\cite{esenturk2018quasi} & 947s & 2603s\\
    \bottomrule
    \end{tabular}
    \parbox{\textwidth}{
    \vspace{1ex} \textit{Note.} This table compares the computational cost of the mloz scheme with the UKCA full chemistry module in UKESM and the Linoz scheme in ICON. For UKESM, the average time cost of mloz and the chemistry module is presented as a percentage of the total runtime of a full-chemistry simulation. The time cost of the UKESM chemistry module includes contributions from chemistry, diagnostics, photolysis, convection, radiation, and dynamics within the UKCA StratTrop mechanism~\cite{esenturk2018quasi}. For ICON, the average wall-clock time for mloz/Linoz is presented in seconds for 1.5-year AMIP-like simulations, measured by the average time differences between Experiment A2/A3 and Experiment A4.}
\end{table}

\section{Summary and discussion}

Over the past two decades, an increasing number of climate models have incorporated complex ozone chemistry and associated chemistry-climate coupling~\cite{IPCC2021}, which comes at a high computational cost. Still, even with today’s computational capacities, comprehensive chemistry schemes remain too expensive for many modeling approaches in climate research, such as large ensembles or convection-permitting resolutions. Around two thirds of the CMIP6 models do not interactively represent ozone changes~\cite{IPCC2021}. In this context, computationally inexpensive ozone schemes that allow adaptive ozone representation are valuable tools. Here, we modeled ozone with a linear ML approach that enables fast, accurate, and stable simulations across a range of climate scenarios. The mloz scheme addresses this need by employing ridge regression to predict daily ozone concentrations based on single-column temperature inputs. We have implemented mloz in two climate models, and evaluated it for two core CMIP experiments in which most climate models still lack interactive ozone representations -- piCTRL and abrupt-4x${\mathrm{CO}_2}$.

Our work establishes the online stability of this first fully integrated ML ozone parameterization within ESMs. We have demonstrated its multi-decadal robustness and accuracy in standard climate sensitivity simulations, and the substantial speed-up compared to full complexity atmospheric chemistry simulations that it has been trained on. The mloz parameterization is around 30 times faster than the UKCA full chemistry module, and around three times faster than Linoz. It produces stable ozone predictions over 50~years under constant forcing, and can respond interactively to a changing environment under CO$_2$ forcing. It well represents various aspects of ozone variability, including seasonal and QBO-related variability, despite a slight underestimation of amplitudes over the stratospheric polar regions. It produces an accurate distribution in stratospheric ozone and total column ozone, with a bias in climatology less than 10\%, despite a slight asymmetry tilting towards the Southern Hemisphere above the tropics. The performance is worse in the troposphere than in the stratosphere due to the more complex ozone chemistry in the troposphere and the daily temporal resolution of mloz, targeting climate-relevant timescales rather than air pollution forecast applications. With mloz we can also realistically emulate the effect of changes in ozone on the modeled climate response under 4x${\mathrm{CO}_2}$ forcing, such as a reduction of upper stratospheric cooling. In that sense, mloz demonstrates strong potential for improving the fidelity of climate change projections, e.g., on stratospheric and tropospheric temperature.

A central advance is that mloz is transferrable across climate models, as we have demonstrated for the successful transfer from UKESM to ICON. In a 30~year-long climate sensitivity simulation, we have tested that mloz produces a stable and highly accurate ozone representation in ICON, compared to its interactive chemistry UKESM ground truth. The bias in ozone climatologies is within 2.5\% in the stratosphere, and within 6\% in the most of the troposphere. We identify two main reasons for the transferability of mloz: first, the vertical temperature profiles in UKESM and ICON are consistent, especially after correcting for the climatological differences in their baseline temperature states during the re-calibration process. Second, the temperature–ozone relationship -- captured by the ridge regression coefficients -- remains stable across both models. These consistencies suggest that mloz can be potentially effectively applied to other climate models as well, since both the vertical temperature structure and the temperature–ozone relationship are governed by fundamental physical and chemical laws that are universally applicable across climate modeling frameworks. Therefore, models lacking a chemistry module can be equipped with a self-consistent ozone representation using mloz parameters trained on another chemistry-climate model.

One might question whether temperature alone can provide sufficient information for predicting ozone across regions. The mloz regression model derives skill from both direct and indirect relationships with temperature. In the upper stratosphere, ozone strongly anti-correlates with temperature due to the temperature dependency of photochemical and catalytic processes~\cite{Hocke2023atmo}. For example, GHG increase (e.g., 4x${\mathrm{CO}_2}$) induces stratospheric cooling, which then leads to significant ozone increase in the tropical mid-upper stratosphere~\cite{Meul2014acp,Chiodo2018jc}. In the lower stratosphere, where ozone has a longer lifetime and is governed primarily by dynamical processes, ozone often manifests in-phase relationship with temperature. For example, a stronger BDC leads to a colder tropical lower stratosphere and enhanced poleward ozone transport~\cite{Randel2021acp}. A stable polar vortex, characterized by cold, isolated polar stratospheric air, hinders poleward transport of ozone across the vortex edge~\cite{Moreira2016acp}. Seasonal and QBO-driven ozone variations are also reflected in temperature variations~\cite{Moreira2016acp,DallaSanta2021jgr}. These processes, though often indirect, are embedded in the temperature field, enabling the model to capture ozone variability across the atmosphere. However, in the troposphere -- where ozone is influenced by emissions, humidity, and convection -- daily-mean temperature alone necessarily performs poorer as a proxy, as reflected in the lower predictive skill of mloz. Including additional variables and increasing the temporal resolution will likely help, although initial online tests did not show major improvements and will also tend to lower the gain in computational efficiency. Expanding the spatial input domain beyond column-wise information improves offline performance, but is difficult to implement online in a portable way due to different parallelization strategies used in climate models. Further research is needed to refine the ML framework and address these technical challenges.

This work opens up new avenues for ML parameterizations in atmospheric chemistry and for a more consistent representation of ozone feedback across climate sensitivity simulations. Future work could for example expand the scheme to more complex scenarios, including interactions with ozone-depleting substances and varying socioeconomic development pathways emission trajectories.

%
%

\section*{Data Availability Statement}
Data archiving is underway. The Fortran code for machine learning implementation in UKESM and ICON, and the Python code for parameterization training, data postprocessing and visualization will be published on Github directory 
\newline(https://github.com/YYilingMa/machine-learning-ozone-parameterization.git) for public access when the paper is accepted. For now the codes are uploaded as Supporting Information for peer-review purpose.

\section*{Competing interests} The authors declare no competing interests.

\section*{Supporting Information} Supporting Information may be found in the online version of this article.

\acknowledgments
The authors acknowledge the research funding provided by the National High-Performance Computing Center at KIT (NHR@KIT), and the computing time provided on the high-performance computer HoreKa at NHR@KIT. This center is jointly supported by the Federal Ministry of Education and Research and the Ministry of Science, Research and the Arts of Baden-W\"urttemberg, as part of the National High-Performance Computing (NHR) joint funding program (https://www.nhr-verein.de/en/our-partners). This work used Monsoon2, a collaborative High-Performance Computing facility funded by the Met Office and the Natural Environment Research Council, and JASMIN, the UK collaborative data analysis facility. This research has been supported by DWD’s “Innovation Programme for Applied Researches and Developments” IAFE ICON-Seamless VH 4.7 (AS). UN got funding from the DFG research unit Volimpact (FOR 2820, grant no. 398006378). We also acknowledge Sven Werchner, Pankaj Kumar and other colleagues at KIT for their technical support in ICON.

%
\bibliography{Reference_arXiv}

\end{document}